\def\authorBlock{
    Youxin Pang\textsuperscript{1} \qquad Ruizhi Shao\textsuperscript{1} \qquad Jiajun Zhang\textsuperscript{2} \qquad Hanzhang Tu\textsuperscript{1} \\ Yun Liu\textsuperscript{1} \qquad Boyao Zhou\textsuperscript{1} \qquad Hongwen Zhang\textsuperscript{3} \qquad Yebin Liu\textsuperscript{1}\footnotemark[1] \qquad \\
    \textsuperscript{1}Tsinghua University \\
    \textsuperscript{2}Beijing University of Posts and Telecommunications\\
    \textsuperscript{3}Beijing Normal University \\
}

\newif\ifreview 
\newif\ifarxiv \newcommand{\arxiv}{\arxivtrue}
\newif\ifcamera 
\newif\ifrebuttal 

\arxiv 

\pdfoutput=1
\documentclass[10pt,twocolumn,letterpaper]{article}
\ifreview \usepackage[review]{cvpr} \fi
\ifarxiv \usepackage[pagenumbers]{cvpr} \fi
\ifrebuttal \usepackage[rebuttal]{cvpr} \fi
\ifcamera \usepackage{cvpr} \fi

\usepackage{graphicx}	
\usepackage{amsmath}	
\usepackage{amssymb}	
\usepackage{booktabs}
\usepackage{times}
\usepackage{microtype}
\usepackage{epsfig}
\usepackage{caption}
\usepackage{float}
\usepackage{placeins}
\usepackage{color, colortbl}
\usepackage{stfloats}
\usepackage{enumitem}
\usepackage{tabularx}
\usepackage{xstring}
\usepackage{multirow}
\usepackage{xspace}
\usepackage{url}
\usepackage{subcaption}
\usepackage{xcolor}
\usepackage[hang,flushmargin]{footmisc}
\usepackage{makecell}

\ifcamera \usepackage[accsupp]{axessibility} \fi

% ...

\ifarxiv  \fi

\newcommand{\R}[1]{{%
    \textbf{%
        \ifstrequal{#1}{1}{\textcolor{red}{R#1}}{%
        \ifstrequal{#1}{2}{\textcolor{blue}{R#1}}{%
        \ifstrequal{#1}{3}{\textcolor{magenta}{R#1}}{%
        \ifstrequal{#1}{4}{\textcolor{teal}{R#1}}{%
                           \textcolor{cyan}{R#1}%
        }}}}%
    }%
}}

\usepackage{xr-hyper}

\makeatletter
\newcommand*{\addFileDependency}[1]{
  \typeout{(#1)}
  \@addtofilelist{#1}
  \IfFileExists{#1}{}{\typeout{No file #1.}}
}

\makeatother
\newcommand*{\myexternaldocument}[1]{
    \externaldocument{#1}
    \addFileDependency{#1.tex}
    \addFileDependency{#1.aux}
}

\definecolor{cvprblue}{rgb}{0.21,0.49,0.74}
\usepackage[pagebackref,breaklinks,colorlinks,allcolors=cvprblue]{hyperref}
\usepackage[capitalize]{cleveref}
\crefname{section}{Sec.}{Secs.}
\crefname{table}{Table}{Tables}
\crefname{figure}{Fig.}{Figs.}

\ifarxiv \crefname{appendix}{App.}{Apps.}
\else \crefname{appendix}{Suppl.}{Suppls.} \fi

\unless\ifarxiv \myexternaldocument{_supplementary} \fi

\begin{document}
\title{ManiVideo: Generating Hand-Object Manipulation Video with Dexterous and Generalizable Grasping}
\author{\authorBlock}

\twocolumn[{
\maketitle
\begin{center}
    \captionsetup{type=figure}
    \includegraphics[width=1.\linewidth]{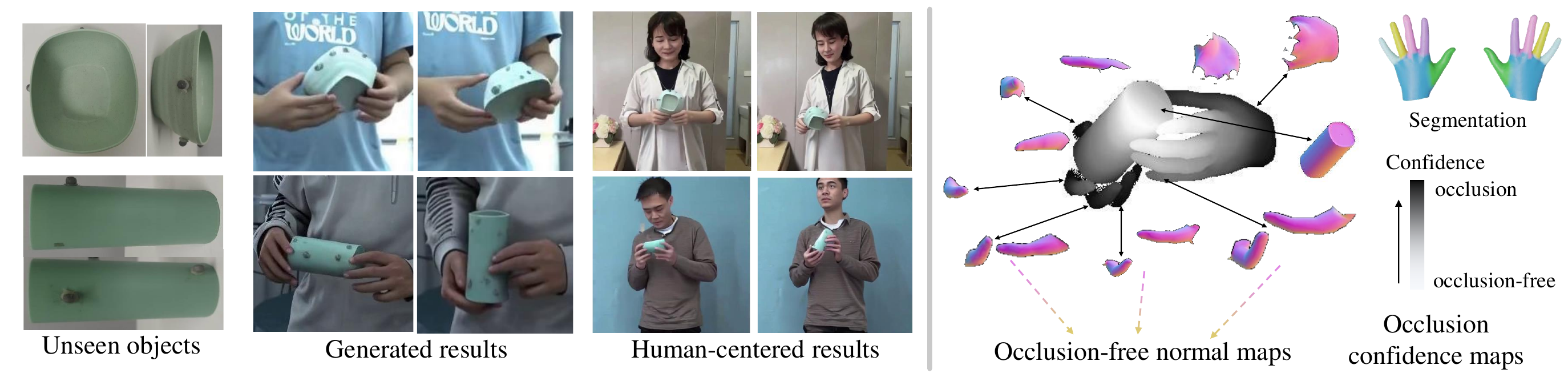}
    \caption{\textbf{ManiVideo}: We propose a novel framework for generalizable and dexterous hand-object manipulation video generation.
    \textbf{Left:} Given several reference images of unseen objects, our method generates realistic and plausible manipulation videos driven by hand-object signals. 
    By integrating multiple datasets, ManiVideo supports applications such as human-centered manipulation video generation.
    \textbf{Right:} To ensure hand-object consistency, we introduce a multi-layer occlusion representation capable of learning 3D occlusion relationships from occlusion-free normal maps and occlusion confidence maps.
}
    \label{fig:insight}
\end{center}
}]

\renewcommand{\thefootnote}{\fnsymbol{footnote}}
\footnotetext[1]{Corresponding Authors.}

\begin{abstract}
In this paper, we introduce ManiVideo, a novel method for generating consistent and temporally coherent bimanual hand-object manipulation videos from given motion sequences of hands and objects. The core idea of ManiVideo is the construction of a multi-layer occlusion (MLO) representation that learns 3D occlusion relationships from occlusion-free normal maps and occlusion confidence maps. By embedding the MLO structure into the UNet in two forms, the model enhances the 3D consistency of dexterous hand-object manipulation. To further achieve the generalizable grasping of objects, we integrate Objaverse, a large-scale 3D object dataset, to address the scarcity of video data, thereby facilitating the learning of extensive object consistency. Additionally, we propose an innovative training strategy that effectively integrates multiple datasets, supporting downstream tasks such as human-centric hand-object manipulation video generation. Through extensive experiments, we demonstrate that our approach not only achieves video generation with plausible hand-object interaction and generalizable objects, but also outperforms existing SOTA methods.


\end{abstract}

\section{Introduction}
\label{sec:intro}

Manipulating objects with bimanual hands is a fundamental component of human activities.
A deep understanding
of hand-object interaction (HOI) during manipulation has been a long-standing challenge in computer vision, playing a crucial role in downstream applications such as HOI reconstruction~\cite{hu2024hand, ye2023diffusion} and virtual reality~\cite{holl2018efficient, wu2020hand}. The key challenge of hand-object manipulation lies in  generating realistic hands and objects while maintaining their natural interactions and temporal consistency.


Recently, methods~\cite{ye2023affordance, hoe2024interactdiffusion, zhang2024hand1000} leveraging the generation capabilities of diffusion models have demonstrated success in generating a large number of realistic HOI images.
A common paradigm in these methods is to leverage geometric representations of hands and objects as conditioning signals to ensure visual plausibility and 3D consistency. Specifically, various forms of structure maps have been explored, including depth maps~\cite{lu2024handrefiner, wang2024rhands}, normal maps~\cite{zhang2024hoidiffusion}, topology maps~\cite{hu2022hand}, hand proxies~\cite{ye2023affordance}, hand masks~\cite{pelykh2024giving}, and bounding boxes~\cite{hoe2024interactdiffusion}, serving as 2D conditions for generating consistent HOI images. 
However, these methods face two main limitations. First, the occlusion relationship between hands and objects is inherently complex. Relying on these 2D conditions only maintain consistency in visible regions while neglecting the plausibility of occluded parts. This limitation poses significant challenges in generating consistent hand-object videos, where complex occlusions naturally arise during the dynamic hand-object manipulation.  Second, due to the limited object diversity in existing HOI datasets used for training, these methods struggle to generalize to diverse objects beyond the training set. 

In this paper, we present a novel framework for generalizable and dexterous hand-object manipulation video generation (ManiVideo), capable of producing realistic, visually compelling videos with 3D consistency and temporal stability, given the motion sequence of hands and objects.
To better understand the occluded regions of HOI, different from existing methods~\cite{hu2022hand, ye2023affordance} that use 2D conditions, we propose to design ManiVideo based on conditional diffusion models and 3D-aware HOI conditions.
Specifically, we introduce a multi-layer occlusion (MLO) representation capable of learning 3D occlusion relationships, which could construct a multi-layerd 3D of hand-object structures and transform them into effective control signals.
As shown on the right side of Fig.~\ref{fig:insight}, MLO representation is composed of two components. 
Inspired by multi-plane images~\cite{tucker2020single}, the first component consists of integral occlusion-free normal maps of HOI, which represents the different layers of hands and objects from far to near.
Each layer indicates a separate finger and is rendered independently to compensate for the hidden regions that are not visible to the 2D signals.
The second are occlusion confidence maps, which can capture the occlusion relationships between visible external areas and hidden internal regions in   occlusion-free normal maps. 
Moreover, to better utilize the MLO structure, we embed it into the denoising UNet~\cite{unet} by concatenating and injecting it into both the initial noise and the added transformer blocks.
Therefore, the model can learn subtle hand-object relationships within the 3D space from our MLO representation, enhancing the 3D consistency of dexterous bimanual hand-object manipulation.

Another challenge in hand-object interaction generation is the scarcity of HOI video data. This limitation restricts the model's ability to generate diverse and realistic objects interactions. A straight-forward way to this limitation is leveraging large-scale 3D dataset like Objaverse~\cite{deitke2023objaverse}, containing over 800K 3D models. However, effectively incorporating these object priors into HOI generation remains challenging. The key lies in bridging two distinct goals: learning generalizable object representations and ensuring their seamless integration with hand interactions.
To address this challenge, we propose a novel paradigm that connects object-only data with hand-object interactions. Our approach first builds rich object priors from Objaverse by rendering multiple viewpoints and simulating basic motion trajectories for each 3D model. We then introduce a novel training strategy that enables simultaneous learning from both object-centric and HOI-centric data. This design leverages the strengths of both sources: HOI videos provide essential interaction patterns, while the abundant object data ensures the diverse object geometry and appearance generation across different time and views. With this novel training strategy, our method achieves generalizable hand-object manipulation and supports additional applications such as human-based hand-object manipulation video generation.

Leveraging the proposed network and training strategies, our method demonstrates the ability to generate hand-object manipulation videos with plausible, dexterous and generalizable grasping.
Through extensive experiments, our method achieves state-of-the-art performance across different datasets.

To summarize, our main contributions are:

\begin{itemize}
    \item The first framework that supports the generation of hand-object manipulation videos with dexterous and generalizable grasping.
    \item A multi-layer occlusion representation that learns articulated occlusions relationship from occlusion-free normal maps and occlusion confidence maps.
    \item A novel training strategy that integrates extensive object datasets and improves the dynamic consistency, thereby enhancing object generalization.
\end{itemize}

\section{Related Work}
\label{sec:related}

\noindent{\textbf{HOI motion generation.}}
Generating plausible hand-object motions plays a crucial role in various downstream tasks such as image/video generation and digital humans. With the introduction of large-scale hand-object interaction datasets~\cite{grab, fan2023arctic, ho3d, liu2024taco, liu2022hoi4d, OakInk, oakink2, obman}, data-driven approaches have achieved considerable advances in synthesizing both static hand grasp~\cite{contactopt, liu2023contactgen, graspingfield, graspTTA, interhandgen} and dynamic grasping~\cite{dgrasp, grip, toch, geneoh} or manipulation~\cite{manipnet, zhang2024manidext, cams, artigrasp, text2hoi}. These high-quality generated or captured hand-object motion sequences can serve as a driving signal for video generation, which in turn provides a more expressive medium for effectively visualizing and presenting interactions.

\noindent{\textbf{HOI image generation.}}
Generating HOI images aids in understanding human activities. 
Current methods fall into three main categories: text-driven, pose-driven, and refinement-based approaches.
Text-driven methods~\cite{narasimhaswamy2024handiffuser, zhang2024hand1000, samuel2024generating} take text prompts as input and use diffusion models to generate realistic images. 
To ensure the consistency of HOI, they rely on additional priors, such as reference images, MANO~\cite{MANO} parameters, and high-dimensional features. 
However, these priors have inherent limitations, as they fail to accurately represent structural information of HOI.
Pose-driven methods~\cite{hu2022hand, ye2023affordance, hoe2024interactdiffusion, zhang2024hoidiffusion, pelykh2024giving, kwon2024graspdiffusion, wang2024realishuman} typically use 2D signals from the hand and object as conditioning inputs, guiding the model to generate HOI images with spatial consistency.
These 2D signals, including normal maps, depth maps, and mask images, offer spatial alignment but fail to capture the full 3D occlusion relationships, such as finger self-occlusion.
HOGAN~\cite{hu2022hand} is closely related to our task.
HOGAN takes a source image and a target pose as inputs, and utilizes flow-based warping functions to transform the source image to align with the target pose.
However, the learned optical flow fails to accurately capture the positional relationship between fingers and objects in 3D space.
Additionally, they are limited to handling only single-handed object interactions.
Refinement-based methods~\cite{lu2024handrefiner, wang2024rhands} aim at refining the rough input hand images to obtain reasonable results.

In addition, GenHeld~\cite{min2024genheld} also leverages objaverse~\cite{deitke2023objaverse} to construct an object code space. 
Specifically, given a hand, GenHeld generates plausible HOI images by retrieving the corresponding object from the object space.
In contrast, our approach generates both hands and objects, focusing on learning appearance and geometric consistency of objects from objaverse.
Cao et al.~\cite{cao2024multi} propose a dual-decoder structure to process hands and objects independently, enabling generalization of both. 
However, their work primarily focuses on generating hand and object motion, with limited exploration of real video generation.

\noindent{\textbf{HOI video generation.}}
HOI video generation remains an unexplored problem due to the limited availability of HOI video data, which makes it challenging to learn generalizable object generation.
Recently, HOI-swap~\cite{xue2024hoi} study the problem of precisely swapping objects in videos. 
While it has contributed to advancing the understanding of HOI tasks in videos, the method is constrained by the simplicity of the interactive actions and the lack of control over the interaction process.
MIMO~\cite{men2024mimo} achieves HOI video generation by decomposing objects and main human. 
However, their focus is primarily on human generation, and they do not thoroughly explore the controllability of HOI.

\noindent{\textbf{Conditional diffusion models.}}
Conditional diffusion models~\cite{zhang2023adding, batzolis2021conditional, ni2023conditional} have driven significant advances in various domains, including image synthesis~\cite{shen2023advancing, dhariwal2021diffusion} and video generation~\cite{wang2024videocomposer, guo2025sparsectrl}. 
Most of them are adapted to large-scale 2D global conditions, such as human skeleton~\cite{hu2024animate}, HED boundary~\cite{zhao2023controlvideo}, and depth maps~\cite{wang2024videocomposer}.
However, due to the dexterity and high degrees of freedom in finger movements, HOI tasks impose more complex structural requirements. 
Compared to 2D global conditions used in most methods, HOI requires more localized and detailed conditions to effectively address the occlusion of hands and objects.

\section{Method}
\label{sec:method}
\begin{figure*}[tp]
    \centering
    \includegraphics[width=1.\linewidth]{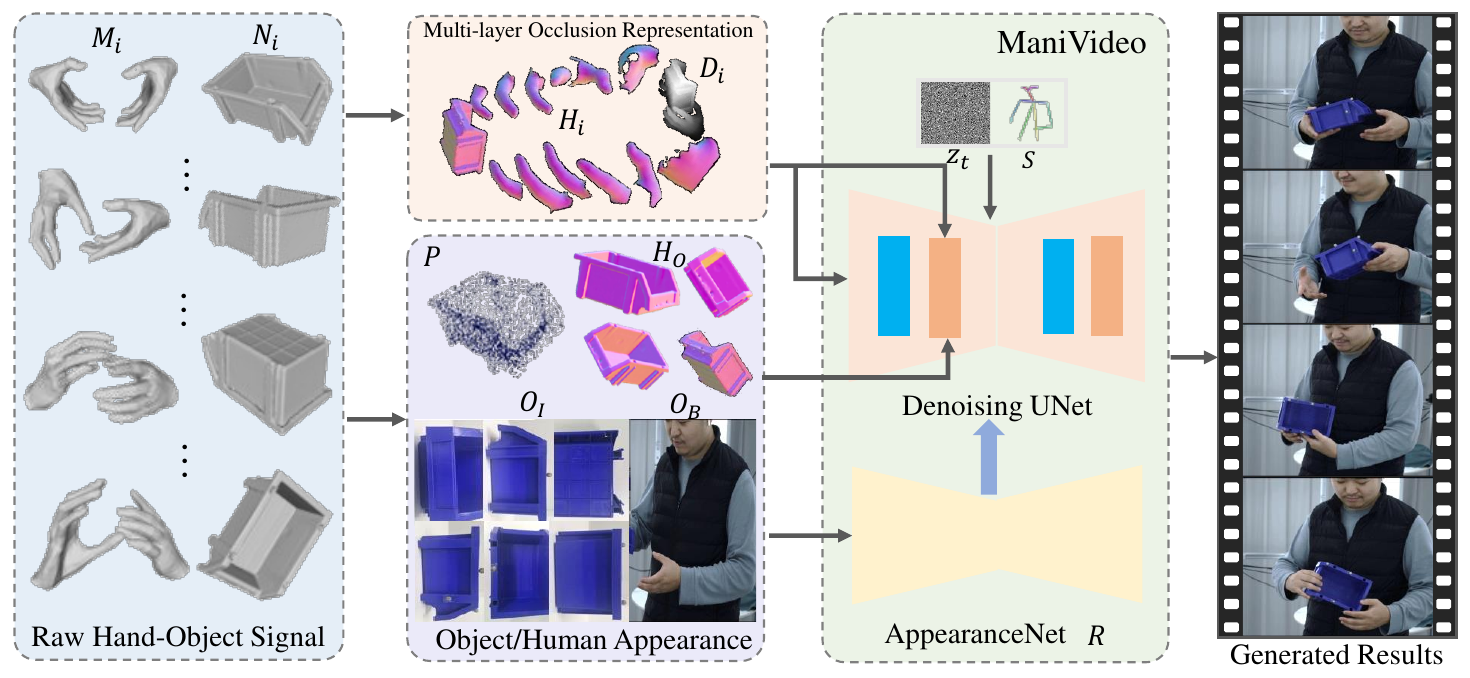}
    \caption{The overall framework of ManiVideo. 
    Given raw hand-object signals, we first transform them into multi-layer occlusion (MLO) representation and object representation.
    MLO structure is designed to enforce the 3D consistency of HOI, which includes occlusion-free normal maps $H$ and occlusion confidence maps $D$.
    Object representation contains the appearance and geometry information, ensuring the dynamic consistency of objects. 
    Then, we inject MLO representation and object representation into the denoising UNet and AppearanceNet.
    }
    \label{fig:network}
\end{figure*}

The overall pipeline of ManiVideo is illustrated in Fig.~\ref{fig:network}.
Given raw hand-object signals of MANO~\cite{MANO} models $M_{i=1}^T = (\theta, \beta)_{i=1}^{T}$ and objects 3D models $N_{i=1}^{T}$, the goal of our method is to generate generalizable and dexterous hand-object manipulation videos.


We first introduce the manipulation conditional representation in Sec.~\ref{subsec: MCR}, including the process of the proposed multi-layer occlusion (MLO) representation (Sec.~\ref{subsec:MAR}) as well as object representation (Sec.~\ref{subsec:ORC}).
The MLO representation is designed to learn occlusion relationships from occlusion-free normal maps $H$, utilizing occlusion confidence maps $D$.
Object representation, including both appearance and geometry, are employed to enhance the 3D consistency of objects.
Then, we explain how the processed conditions are embedded into the proposed ManiVideo in Sec.~\ref{subsec: MANI}.
After that, we introduce designed training strategies that can effectively integrate multiple datasets and address downstream tasks, including human-centered hand-object manipulation video generation, by fine-tuning ManiVideo with human data (Sec.~\ref{subsec: TS}).





\subsection{Manipulation Conditional Representation}~\label{subsec: MCR}
In this section, we introduce the proposed MLO representation and object representation.
\subsubsection{Multi-layer Occlusion Representation}~\label{subsec:MAR}
HOI generation requires precise handling of complex occlusions and articulations in 3D space, including both self-occlusions and mutual occlusions.
Directly embedding raw hand-object signals of $M$ and $N$ (the left side of Fig.~\ref{fig:network}) into the UNet can lead to artifacts and penetration (as shown in Fig.~\ref{fig:exp_abl}), as the 2D conditions used in \cite{ye2023affordance, zhang2024hoidiffusion} are unable to adequately account for the hidden regions.
To address this, we propose a multi-layer occlusion (MLO) representation (the second column in Fig.~\ref{fig:network}), which includes occlusion-free normal maps to provide integral hand-object signals and occlusion confidence maps to indicate the degree of occlusion.

Specifically, we organize hands and objects into occlusion-free normal maps $H_{i=1}^{T} = {\phi(N_{i}^{T}, M_{i}^{T})}$, which includes the object, the palm, the thumb, the index finger, the middle finger, the ring finger, and the little finger.
$\phi$ represents the process of projecting 3D space into 2D image space using specific camera parameters.
For hand segmentation, we employ the hand semantic map~\cite{liu2023contactgen} to independently render the normal map for each part, thereby compensating for the limitations of 2D signals that overlook occluded areas.

To learn the occlusion relationship between visible and hidden regions in 3D space, we utilize depth maps as occlusion confidence maps $D_{i=1}^{T}$ to represent the degree of occlusion in different parts.
As shown in Fig.~\ref{fig:insight}, the relatively darker the region in the confidence map, the more severe the occlusion in the corresponding part.
Thus, the model can distinguish between visible and hidden regions using $D$, allowing it to retain occlusion-free parts while effectively refining the occluded parts based on $H$.
Furthermore, our method is capable of improving results in bimanual hand manipulation generation, as the MLO representation is designed to simultaneously represent bimanual hands and objects.
Besides, to fully capitalize on the advantages of MLO structure, we incorporate it into the UNet in two distinct ways (Sec.~\ref{subsec: EM}).

\subsubsection{Object Representation}~\label{subsec:ORC}
Another challenge is ensuring object generalization, particularly given the limited availability of HOI video data. 
Most existing datasets comprise only over 10 object categories~\cite{liu2022hoi4d, chao2021dexycb, hampali2021ho}, which restricts the model to learn the correspondence between appearance and geometry.
Therefore, integrating more object information into the model is crucial.
The recent release of Objaverse~\cite{deitke2023objaverse} has significantly enhanced the diversity and scale of 3D repositories. 
To achieve 3D consistent object generation, we incorporate Objaverse into our training dataset.

Specifically, for each object in Objaverse, we utilize the base mesh as front reference and obtain appearance images ${O_{I}}$ from six distinct viewpoints: front, back, left, right, top, and bottom (as shown in the second column of Fig.~\ref{fig:network}). 
These reference images aim to ensure appearance consistency during movement. 
Additionally, we incorporate a human/background reference image $O_{B}$ to provide the generated scene.
${O_{I}}$ and $O_{B}$ are then embedded into UNet through AppearanceNet $R$ (Sec.~\ref{subsec: BOT}).

To accurately simulate object motions in HOI video data, we randomly generate a sequence of quaternion $Q_{i=1}^{T} = (a, b, c, w)_{i=1}^{T}$ and translation $L_{i=1}^{T} = (x, y, k)_{i=1}^{T}$ for each object, where $Q$ denotes the object's rotation and $L$ represents its position. 
The object's model $N$ and its corresponding motion sequence ($Q$, $L$) are combined to render the normal maps $H_{o}=\phi(N, Q, L)$, where $\phi$ represents projecting 3D mesh into 2D images.
Additionally, to reinforce structural constraints, we uniformly sample the mesh surface to generate point cloud $P \in \mathbf{R}^{2048 \times 3}$.

\subsection{ManiVideo}~\label{subsec: MANI}
Inspired by Animate Anyone~\cite{hu2024animate}, our model comprises an AppearanceNet $R$ and a denoising UNet, as shown in the green box of Fig.~\ref{fig:network}.
$R$ is responsible for appearance features of object and background, while the UNet generates the results by denoising the input noisy latent $z_{t}$.

\subsubsection{Embedding of MLO Representation}~\label{subsec: EM}
To enhance 3D consistency, two ways are employed to embed $H$ and $D$ into the diffusion UNet.
Firstly, to make the results align precisely with the given conditions, 
we utilize the spatial correspondence between $H$ and the output to directly impose structural constraints on the initial noise.
Specifically, similar to \cite{hu2024animate}, we employ a lightweight pose guider, which includes four convolution layers, to align $H$ with the same resolution as the noise latent $z_{t}$.
The cascaded $H$ along the channel is passed to the pose guider (G) to extract features, which are then injected by adding them to $z_{t}$:
\begin{equation}
    z_{t}^{'} = z_{t} + G([H])
\end{equation}

Secondly, to extract the 3D spatial occlusion relationships embedded in MLO structure, we also integrate $H$ and $D$ into the added transformer blocks.
Specifically, we concatenate $H$ and $D$, and extract the embedding $E_{F}$ through convolution layers and MLPs. 
Then $E_{F}$ is injected into transformer blocks using cross attention $\operatorname{Attention}(Q, K, V)=\operatorname{Softmax}\left(\frac{Q K^T}{\sqrt{d}}\right) \cdot V $:
\begin{equation}
    \begin{aligned}
	    E_{F} &= MLP(Conv[H, D]))\\
        Q=&W^Q z, K=W^K E_{F}, V=W^V E_{F} \\
    \end{aligned}
    \label{eq:two}
\end{equation}
where $z$ is the feature from previous blocks, $W^Q$, $W^K$ and $W^V$ are learnable matrices that project the inputs to query, key and value, respectively, while $d$ represents the output dimension of key and query features.
Eq.~\ref{eq:two} encourages the model to leverage $H$ and $D$ to capture 3D occlusion relationships.
These two distinct embeddings ensure that the model learns coarse spatial correspondences in the initial layers and perceives complex occlusion relationships in the deeper layers.

\subsubsection{Embedding of Object Representation}~\label{subsec: BOT}
To independently process appearance and geometry of object into UNet, we embed them into different places. 

Specifically, we incorporate the appearance information into the UNet via $R$.
Given the feature $f_{0} \in \mathbb{R}^{(b \times t) \times h \times w \times c}$ obtained from the denoising UNet, we first utilize $R$ to extract the features $f_{i=1}^{7} \in \mathbb{R}^{b \times h \times w \times c}$ from reference images $(O_{I}, O_{B})$.
Subsequently, we repeat $f_{i=0}^{7}$ by $t$ times and concatenate it along the width channel.
The concatenated feature is then adjusted to match the dimensions of $f_{0}$ through convolution operations, before sending $f^{'}$ to the next block of the denoising UNet:
\begin{equation}
    f^{'} = Conv([R([O_{I}, O_{B}]), f_{0}])
    \label{eq:one}
\end{equation}
Optionally, to establish the initial state of the object, we can additionally provide the object image of the first frame and extract features as $f_{8}$.



Furthermore, to enable the model to capture geometry consistency, we first utilize convolution layers and MLPs to extract the geometry embedding $E_{N}$ of $H_{o}$ and $P$, which is then injected into added transformer blocks via cross attention $\operatorname{Attention}(Q, K, V)=\operatorname{Softmax}\left(\frac{Q K^T}{\sqrt{d}}\right) \cdot V $:
\begin{equation}
    \begin{aligned}
	    E_{N} &= MLP(Conv[H_{o}, P]))\\
        Q=&W^Q z, K=W^K E_{N}, V=W^V E_{N} \\
    \end{aligned}
\end{equation}
This approach enables the utilization of extensive dynamic 3D object data to improve dynamic consistency in appearance and geometry, effectively compensating for the scarcity of HOI video data.

\subsection{Training Strategy}~\label{subsec: TS}
To connects object-only data with hand-object interactions, we propose novel training strategies that utilize all dataset effectively.
The data used by ManiVideo comprises a limited set of HOI video data, alongside a substantial collection of synthetic object-only data.
The complete training process consists of two stages: the image stage and the temporal stage, which are trained independently in sequence.

In the image stage, we utilize the pipeline described in Sec.~\ref{subsec: MCR} to process the MLO structure and object representation for HOI video data.
For Objaverse, we only simulate the movement of objects and ignore hands.
We set hand-related layers in MLO structure to zero, while keep object representation unchanged.
These two conditions are then embedded into unet with different ways, corresponding to Sec.~\ref{subsec: EM} and Sec.~\ref{subsec: BOT} respectively.
In the temporal stage, we freeze the parameters at the image stage and add temporal layers like Animate Anyone~\cite{hu2024animate} for independent training.
Through the above two stages, the model gains the ability to generate visually appealing results from HOI video data while also enhancing object generalization by learning dynamic consistency from Objaverse.

Additionally, to enable downstream tasks such as human-based HOI video generation in Sec.~\ref{sub:app}, ManiVideo can be fine-tuned with human-centered datasets, including Tiktok~\cite{jafarian2021learning} and Human4DiT~\cite{shao2024human4dit}.
Since human videos lack object information, and hand regions are often small or affected by motion blur, we use a single human image as the background reference image $O_{B}$, setting all other conditions to zero.
Optionally, to enhance the control of the human body, we extract the human skeleton $S$ and utilize the pose guider $G_{1}$ with the same structure as $G$ to capture relevant features. 
Finally, features are added to the noisy input:
\begin{equation}
    z_{t}^{''} = z_{t} + G_{1}(S)
\end{equation}
In this way, our method allows for the integration of three distinct datasets, enabling the ManiVideo to jointly generate the human, objects, and hands.

\section{Experiments}
\label{sec:expe}

\subsection{Settings}
\noindent{\textbf{Dataset.}}
We train our ManiVideo on three types of datasets.
For object data, we utilize Objaverse~\cite{deitke2023objaverse}, and human data is sourced from Human4DiT~\cite{shao2024human4dit}.
For HOI video data, in addition to the public DexYCB dataset~\cite{chao2021dexycb}, We collect third-person view videos of participants standing and interacting with objects using bimanual hands.
Specifically, our dataset contains 722 videos (376k frames) depicting daily tool-use behaviors covering 15 objects, 10 views and 8 participants.
Compared to other datasets~\cite{liu2024taco, fan2023arctic}, our data is human-centric and free from distractions caused by irrelevant objects (e.g., tables), making it particularly suitable for downstream applications like human-based HOI video generation in~Sec.\ref{sub:app}.
For each data, we select 3\% as the test set.
 
\begin{figure}[tp]
    \centering
    \includegraphics[width=1.\linewidth]{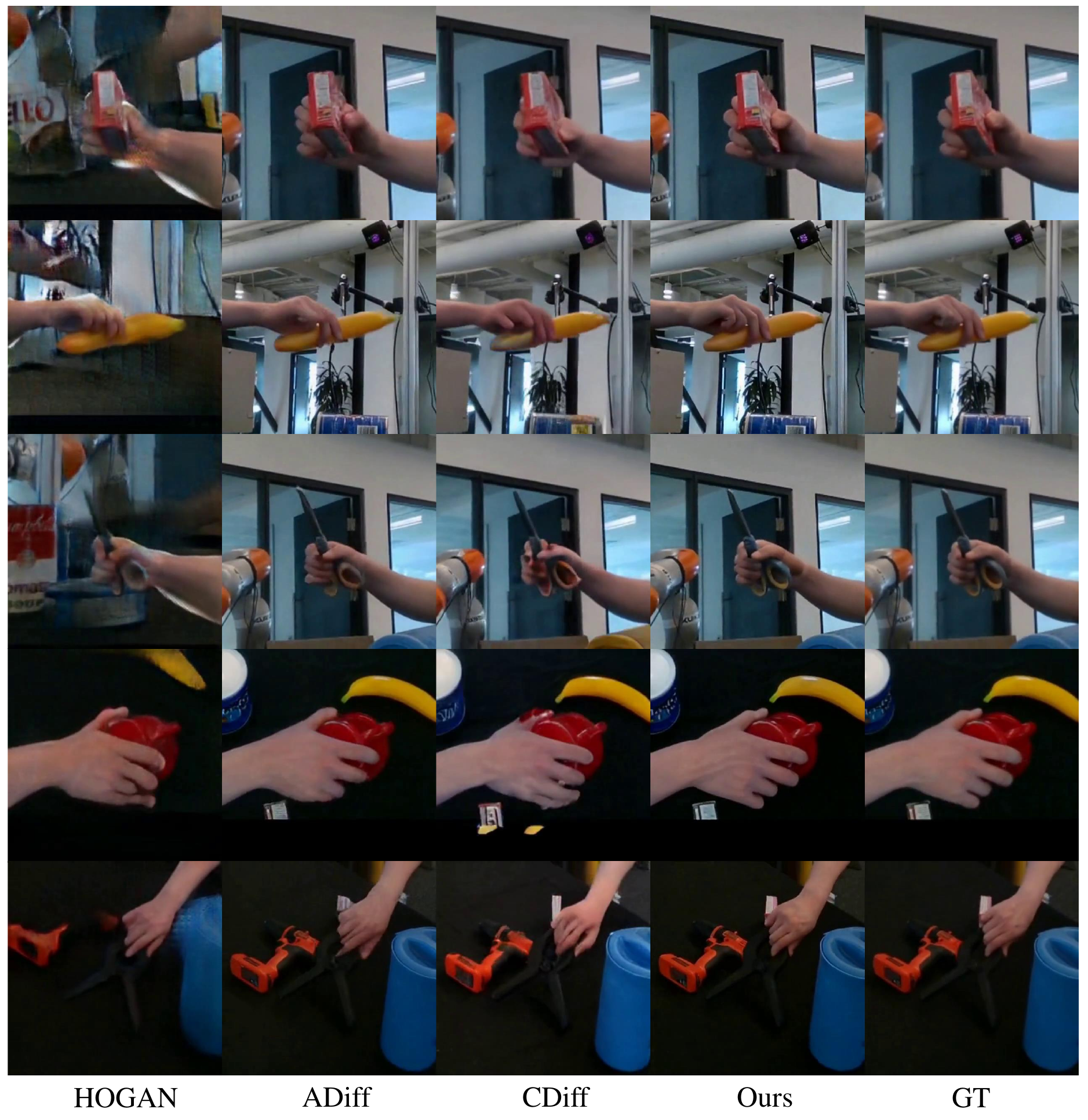}
    \caption{Qualitative comparison of different methods on DexYCB dataset~\cite{chao2021dexycb}. Our results perform best in cases of hand-object mutual occlusion and finger self-occlusion.}
    \label{fig:exp_fig1}
\end{figure}

\noindent{\textbf{Metrics.}}
To evaluate generation quality, we adopt the widely used Fréchet Inception Distance (FID)~\cite{dowson1982frechet}, Peak Signal-to-Noise Ratio (PSNR), Structural Similarity Index (SSIM), and Learned Perceptual Similarity (LPIPS)~\cite{zhang2018unreasonable}.
To evaluate the accuracy of finger generation, we use the Mean Per-Joint Position Error (MPJPE) as employed in~\cite{fan2023arctic}, which measures the L2 distance between the 21 predicted hand joints and the ground truth joints.

\begin{figure}[tp]
    \centering
    \includegraphics[width=1.\linewidth]{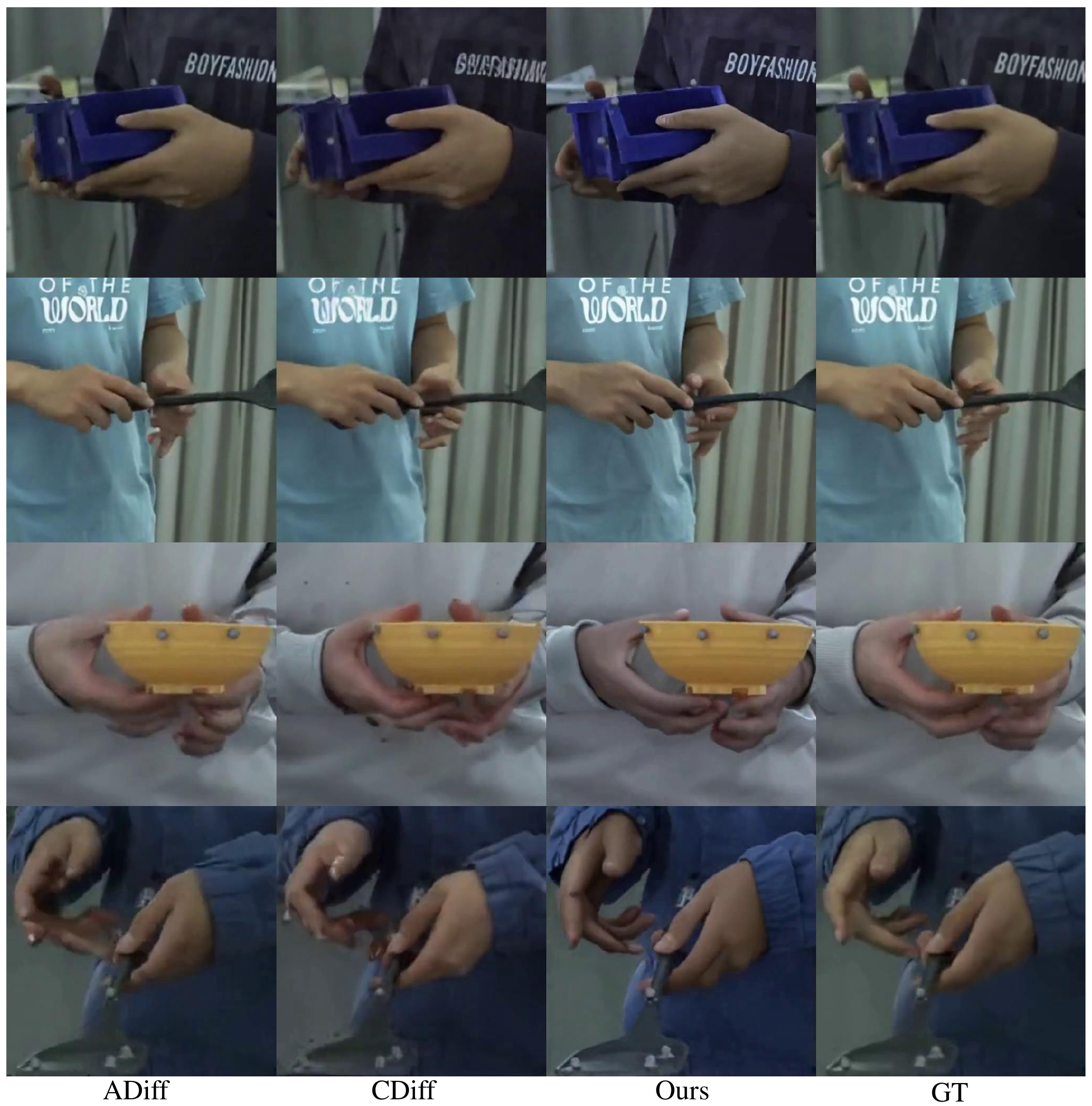}
    \caption{Qualitative comparison of different methods on videos we collect. Our approach achieves the best results.}
    \label{fig:exp_fig2}
\end{figure}

\noindent{\textbf{Implementation Details.}}
We train ManiVideo in a two-stage manner, sampling equally from object, HOI, and human datasets in each iteration.
During the first stage, non-temporal layers are optimized over approximately 20,000 iterations, followed by temporal layer optimization for around 30,000 iterations. 
The batch size is set to 4 in the first stage, reducing to 1 in the second stage, where we use 24-frame video sequences.
Adam \cite{kingma2015adam} is selected as the optimizer with a learning rate of $1e-5$ for the first stage and $8e-6$ for the second one. 
All experiments are conducted on a single A800 GPU.

\subsection{Baseline Comparisons}
In this part, we mainly verify the plausibility and consistency of HOI generation.
We compare our method with three others: two focusing on HOI generation and one conditional method based on ControlNet~\cite{zhang2023adding}.
Specifically, Hand-Object Interaction Image Generation (HOGAN)~\cite{hu2022hand} is a GAN-based approach that takes source images and target poses as inputs to generate corresponding results.
HOGAN constructs a unified surface space with a model-aware hand-object representation, independently generating the background, objects, and hands.
Affordance Diffusion (ADiff)~\cite{ye2023affordance} aims to synthesize plausible images of hand-object interactions given single object reference image.
In addition, to emphasize the effectiveness of the proposed MLO structure and embedding, we also utilize ControlNet combined with Stable Diffusion v1.5~\cite{rombach2022high} (CDiff) to achieve the generation of HOI images.

\begin{table}[t]
   \centering
   \small
   \begin{tabular*}{1.\linewidth}{c|ccccc}
  \hline
  \multirow{2}*{Method}   & \multicolumn{5}{c}{\makecell[c]{DexYCB \\ }}\\
    & FID$\downarrow$ & LPIPS$\downarrow$ & PSNR$\uparrow$ & SSIM$\uparrow$ & MPJPE$\downarrow$\\ \hline
       \makecell[c]{HOGAN}  &  64.74 & 0.102 & 29.50 &  0.896 & 60.95 \\ 
       \makecell[c]{ADiff}  & 53.95 & 0.093 & 29.96 & 0.903 &  59.12\\ 
       \makecell[c]{CDiff} & 84.74 & 0.127 & 28.27 & 0.835 & 68.01\\ 
       \makecell[c]{Ours}  & \textbf{49.96} & \textbf{0.079} & \textbf{30.10} & \textbf{0.913} & \textbf{57.30}\\
       \hline
  \multirow{2}*{Method}   & \multicolumn{5}{c}{\makecell[c]{Our Dataset\\ }}\\
    & FID$\downarrow$ & LPIPS$\downarrow$ & PSNR$\uparrow$ & SSIM$\uparrow$ & MPJPE$\downarrow$\\ \hline
       \makecell[c]{ADiff}  &  39.91 & 0.127 & 29.17 & 0.898 & 37.45 \\ 
       \makecell[c]{CDiff} &  45.50 & 0.133 & 28.33 & 0.883 &  42.89 \\ 
       \makecell[c]{Ours}  & \textbf{37.70} & \textbf{0.113} & \textbf{29.59} & \textbf{0.905} & \textbf{32.89} \\
       \hline
  \end{tabular*}
  \vspace{5pt}
   \caption{Quantitative comparison on DexYCB and our dataset.
   Our ManiVideo outperforms other methods.}
   
   \label{tab:exp_tab1}
\end{table}
Moreover, to ensure a fair comparison, we apply specific adjustments to each method.
For HOGAN, since no pre-trained model is provided, we train it from scratch.
Notably, as HOGAN is designed for single-handed object interactions,
we only evaluate it on the DexYCB dataset~\cite{chao2021dexycb} following the official instructions.
For ADiff, we only use ContentNet and replace the original input mask with the depth maps of hands and objects, while using single object image as reference.
For CDiff, similar to ADiff, we use the object reference image and HOI depth as conditioning inputs, and fine-tune both SD1.5 and ControlNet concurrently.
In our method, we incorporate the MLO representation and object representation as inputs (Sec.~\ref{sec:method}) to generate the corresponding results.

\begin{figure}[tp]
    \centering
    \includegraphics[width=1.\linewidth]{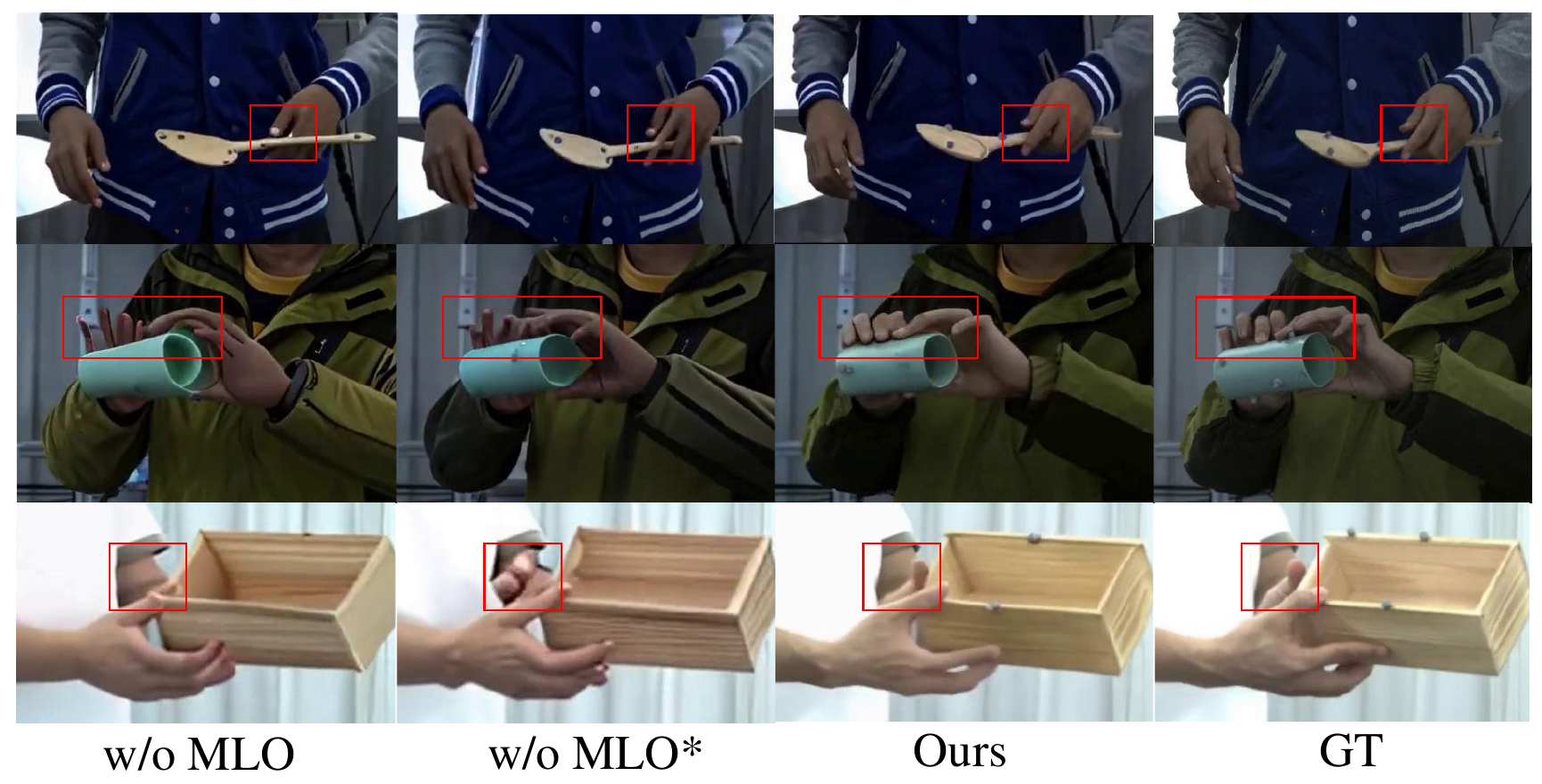}
    \caption{Ablation study of the multi-layer occlusion (MLO) representation. Without MLO structure, basic 2D conditions fail to ensure accurate structure and occlusion relationships between objects and fingers.
    Incomplete embedding (w/o MLO*) diminishes the effectiveness of the MLO representation.}
    \label{fig:exp_abl}
\end{figure}

\noindent{\textbf{Qualitative Evaluation.}}
The visual comparisons for the DexYCB dataset~\cite{chao2021dexycb} are shown in Fig.~\ref{fig:exp_fig1}.
Since HOGAN uses background inpainting and fuses it with the generated hands and objects, the generated background differs from other methods.
The analysis is summarized as follows.
Our approach achieves superior dynamic finger generation compared to other methods, particularly in scenarios involving finger self-occlusion.
For instance, as illustrated in the first and third rows, several fingers are positioned closely, with parts of the hand occluded by the object.
The results produced by HOGAN and CDiff display noticeable artifacts, while ADiff's output contains an incorrect number of fingers.
In the second row, overlapped fingers result in poor generation for the occluded fingers by both ADiff and CDiff.
In the last two rows, HOGAN, ADiff, and CDiff fail to produce accurate results due to the invisibility of bent fingers.
Specifically, directly learning the hand-object correspondence from 2D conditions is an ill-posed problem, which prevents HOGAN from effectively capturing consistency when the fingers are densely packed.
For diffusion-based methods, the relatively small proportion of fingers can lead to details being misinterpreted during denoising process. 
Conversely, our method utilizes the proposed multi-layer occlusion representation to effectively learn occlusion relationships from integral finger information. 
This approach addresses challenges such as finger self-occlusion, mutual occlusion between hands and objects, and the invisibility of bent fingers, enabling more accurate handling of these complex interactions.

The experimental results on our collected dataset are shown in Fig.~\ref{fig:exp_fig2}. As HOGAN is limited to handling single-hand interactions, we exclude it from comparison.
The figure primarily demonstrates the mutual occlusion between objects and hands, as well as the self-occlusion among fingers. 
Although both ADiff~\cite{ye2023affordance} and CDiff~\cite{zhang2023adding} are diffusion-based methods, the dexterity and high degrees of freedom of the fingers make it challenging to generate reasonable results when directly conditioned on 2D inputs.
On the contrary, our MLO representation plays an important role in helping the model perceive 3D relationships.

\begin{figure}[tp]
    \centering
    \includegraphics[width=1.\linewidth]{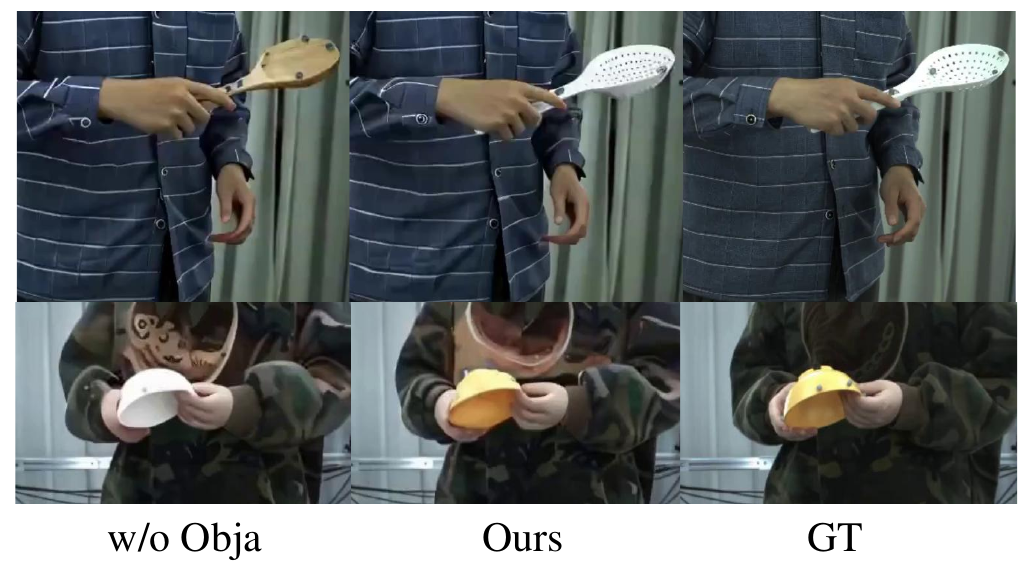}
    \caption{Ablation study of object augmentation training. 
    Utilizing Objaverse helps the model learn dynamic consistency from large object datasets.}
    \label{fig:exp_abl2}
\end{figure}

\noindent{\textbf{Quantitative Evaluation.}}
The quantitative comparisons of different methods are shown in Tab.~\ref{tab:exp_tab1}.
Since different methods handle the background in various ways, we calculate metrics only for the hand-object areas.
It can be observed that our method outperforms others in terms of fidelity and 3D consistency. 
These quantitative results align with the observation in the visual results.

\subsection{Ablation Studies}

\textbf{Multi-layer Occlusion Representation}.
First, we demonstrate the effectiveness of the proposed MLO representation.
To improve the model's ability to perceive both visible and occluded regions, we introduce MLO structure, which learns 3D consistency using integral hand-object information.
As shown in Fig.~\ref{fig:exp_abl}, w/o MLO indicates injecting only depth maps into UNet, while w/o MLO* injects MLO structure solely into the initial noise.
In the first row, the absence of the MLO structure limits the model's capacity to capture occlusion relationships, while incomplete embedding can result in penetration.
In the second row, the hand-object relationship is misinterpreted when fingers make contact with the object.
The third row shows that severe occlusions between hands and objects result in noticeable artifacts in occluded regions for both w/o MLO and w/o MLO*. 
Therefore, MLO representation provides a more integral 3D perspective, and complete embedding effectively utilizes this information.

\begin{figure}[tp]
    \centering
    \includegraphics[width=1.\linewidth]{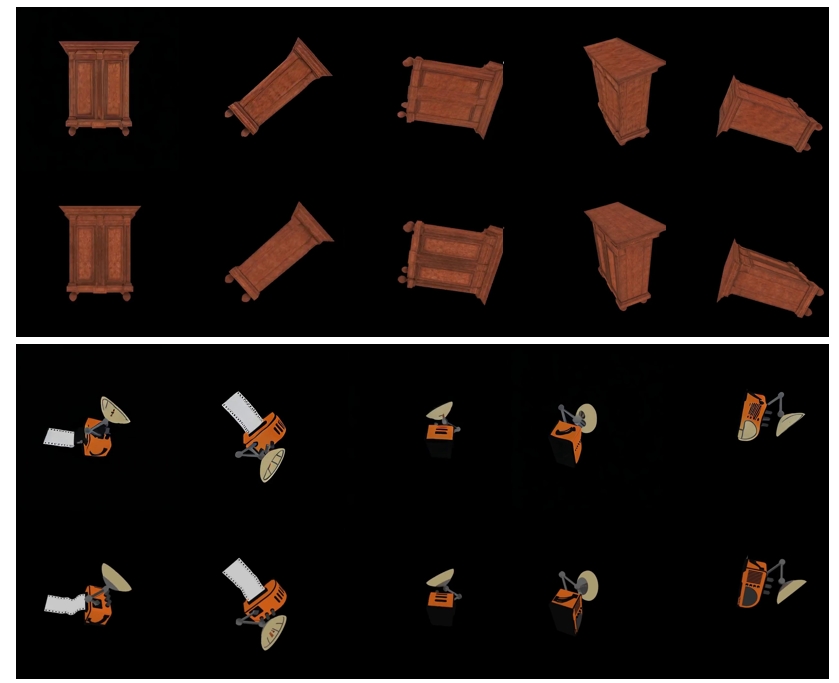}
    \caption{Results on objaverse~\cite{deitke2023objaverse}. For each example, the first row shows the generated results, while the second row is the ground truth. The results show that our model learns the consistency of objects from Objaverse.}
    \label{fig:obja}
\end{figure}

\begin{table}[t]
   \centering
   \small
   \begin{tabular*}{1.\linewidth}{c|ccccc}
  \hline
  Method & FID$\downarrow$ & LPIPS$\downarrow$ & PSNR$\uparrow$ & SSIM$\uparrow$ & MPJPE$\downarrow$\\ \hline
       w/o Obja  & 61.60 & 0.121 & 27.99 & 0.895 & 37.33\\ 
       w/o MLO  & 46.67 & 0.115 & 28.26 & 0.869 & 39.41\\ 
       w/o MLO* & 40.60 & 0.117 & 28.30 & 0.881 & 34.02\\ 
       Ours  & \textbf{37.70} & \textbf{0.113} & \textbf{29.59} & \textbf{0.905} & \textbf{32.89}\\
       \hline
  \end{tabular*}
  \vspace{5pt}
   \caption{Quantitative comparison for ablation studies.
   Complete ManiVideo achieves the optimal result.}
    
\vspace{-6mm}
   \label{tab:exp_tab2}
\end{table}
\noindent{\textbf{Embedding of Objaverse.}}
To validate the effectiveness of objaverse embedding, we exclude Objaverse~\cite{deitke2023objaverse} and train ManiVideo only with our collected video dataset. 
As shown in Fig.~\ref{fig:exp_abl2}, due to the limited HOI video data, although the object structure aligns with the MLO representation, the model tends to overfit dynamic texture details from the training data, making the condition of object reference images less effective.
Therefore, it is crucial to leverage large-scale 3D object data to enhance the model consistency in both appearance and geometry.

Furthermore, the effectiveness of objaverse embedding is highlighted by direct demonstration of the object generation results in Fig.~\ref{fig:obja}.
Specifically, we randomly select two objects from objaverse as the test set. 
For each object, we construct object representation and MLO structure as inputs, as described in Sec.~\ref{subsec: TS}.
Fig.~\ref{fig:obja} illustrates the appearance retention capability of dynamic objects, with each row depicting the same object with different motion states.

Quantitative comparisons using the two modules are shown in Tab.~\ref{tab:exp_tab2}.
It can be seen that the complete framework achieves the optimal results. 
Omitting MLO representation and objaverse embedding leads to performance degradation.

\subsection{Applications}   \label{sub:app}

\begin{figure}[tp]
    \centering
    \includegraphics[width=1.\linewidth]{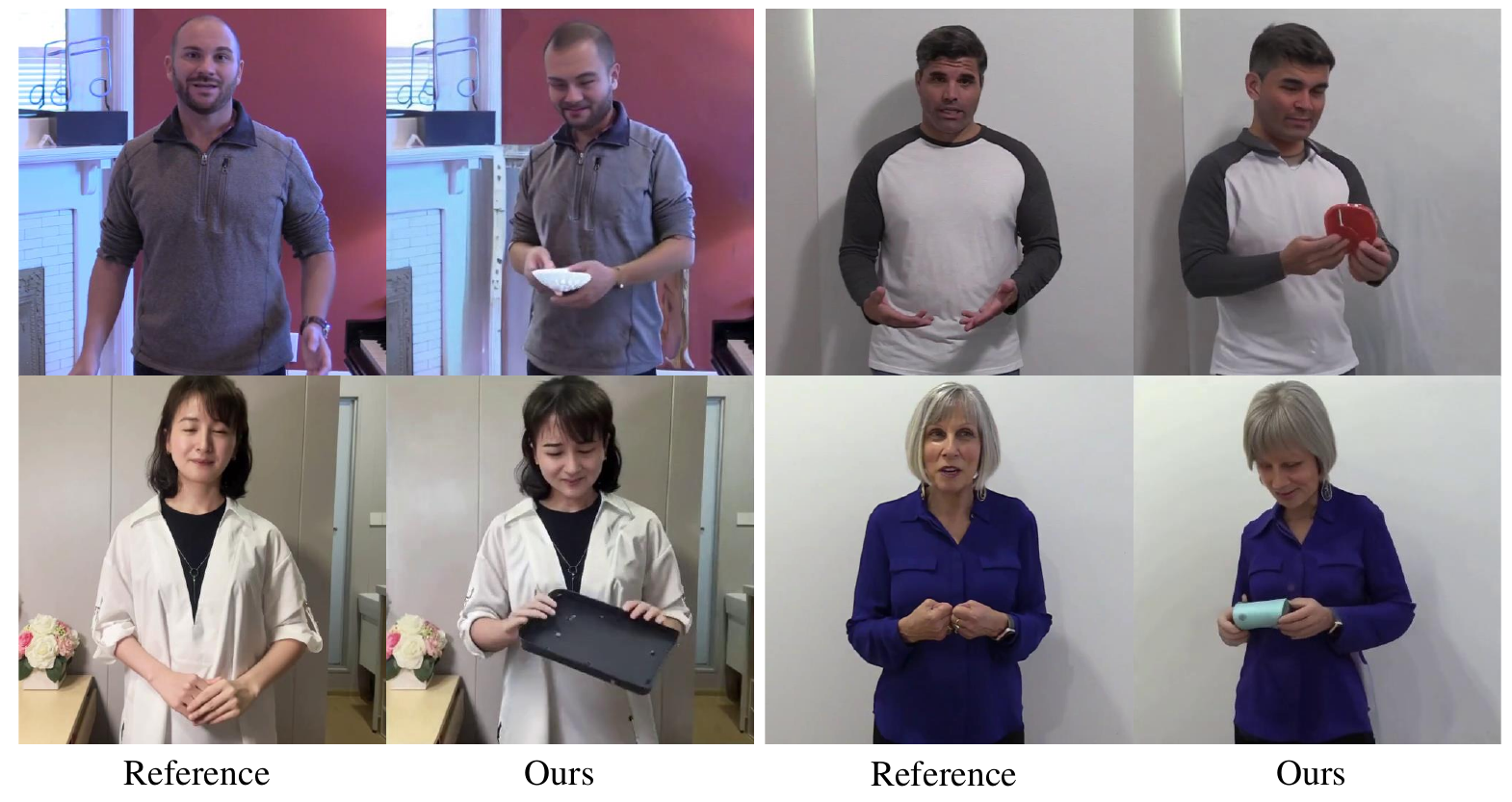}
    \caption{Human-based hand-Object manipulation video generation. 
    Using human reference images as input and training on human-centered datasets, our model is capable of generating human-centric hand-object manipulation videos.}
    \label{fig:app}
\end{figure}

\noindent{\textbf{Human-based HOI video generation.}}
In Fig.~\ref{fig:app}, we show the results of incorporating a single human into hand-object manipulation video generation, which provide support for handheld object digital humans and human-object interactions tasks.
By incorporating human data~\cite{shao2024human4dit} as one of the training sets, ManiVideo can effectively learn HOI while maintaining generalization across various human identities.
Given a human-centered reference image, we use it as the human/background reference image $O_{B}$, embedding it into UNet through $R$.  
Optionally, to ensure the stability of human bodies, the human skeleton can be used as driving signals and embedded into the UNet (Sec.~\ref{subsec: TS}).


\section{Conclusion}
\label{sec:conclusion}

In this work, we proposed ManiVideo, an innovative framework for generating temporally consistent and realistic bimanual hand-object manipulation videos based on motion sequences. 
Our approach proposes a multi-layer occlusion (MLO) representation, enabling the model to effectively learn complex 3D occlusion relationships through both occlusion-free normal maps and occlusion confidence maps. 
To address the challenge of limited video data and to enhance the model‘s ability to generalize across various objects, we utilize Objaverse, a large-scale 3D object dataset, into the training process, leading to richer object consistency learning. Furthermore, we design a novel training strategy that allows the simultaneous use of multiple datasets, further supporting tasks such as human-centered hand-object manipulation video generation. 

\noindent{\textbf{Limitation.}}
The performance of ManiVideo is constrained by the accuracy of the driving signals. 
Moreover, the generalization of complex object textures is still constrained by the domain gap between synthetic and realistic data. 
Incorporating the object appearance to viewpoint and time through more meticulous 4D representation could improve this.


{\small
\bibliographystyle{ieeenat_fullname}
\bibliography{11_references}
}


\end{document}